\documentclass{article}
\usepackage{spconf,amsmath,graphicx}
\usepackage{color}
\usepackage{multirow}

\newcommand{\COMMENT}[1]{}

\title{NON-NATIVE CHILDREN SPEECH RECOGNITION THROUGH TRANSFER LEARNING}
\name{Marco Matassoni, Roberto Gretter, Daniele Falavigna,  Diego Giuliani}
\address{Center for Information and Communication Technology\\Fondazione Bruno Kessler, via Sommarive 18, Trento (Italy) \\
{\small \tt{\{matasso,gretter,falavi,giuliani\}@fbk.eu}}}
\begin{document}
%\ninept
%
\maketitle
\begin{abstract}
This work deals with non-native children's speech and investigates both multi-task and transfer learning approaches to adapt a multi-language Deep Neural Network (DNN) to speakers, specifically children,  learning a foreign language. 
The application scenario is characterized by young students learning English and German and reading sentences in these second-languages, as well as in their mother language. The paper analyzes and discusses techniques for training effective DNN-based acoustic models starting from children native speech and performing adaptation with limited non-native audio material. 
A multi-lingual model is adopted as baseline, where a common phonetic lexicon,  defined in terms of the units of the International Phonetic Alphabet (IPA), is shared across the three languages at hand (Italian, German and English); DNN adaptation methods based on transfer learning are evaluated on significant non-native evaluation sets.    
Results show that the resulting non-native models allow a significant improvement with respect to a mono-lingual system adapted to speakers of the target language.    
\end{abstract}
\section{Introduction}
\label{sec:intro}

Nowadays the usage of deep neural networks hidden Markov models (DNN-HMMs) \cite{hinton2012,mohamed2012}  provides effective performance in speech recognition: there are  
concrete applications ranging from mobile voice search \cite{yao2013}, transcriptions of broadcast news, videos \cite{dahl2012} or conversations  \cite{telseide2011} to recognition in noisy environments \cite{seltzer2013,barker2015,renals2014}. 

The availability of large training corpora for a given application domain allows to train a DNN with many layers and parameters in order to improve the classification performance. On the contrary, in the absence of sufficient data for training, e.g.\ in the case of under-resourced languages, the number of DNNs parameters that can be reliably estimated greatly reduces and, consequently, classification performance is not always satisfactory. Recognition of children's speech is a kind of application domain often characterized by training data shortage, even for major languages. 

As alternative to complete training of DNN parameters, starting by scratch, adaptation of  an existing DNN by using the available small data set is a viable approach. This has been investigated in \cite{Matassoni_SLT16,serizel2016}, where an initial DNN trained on adult speakers is then adapted using limited set of children's data.

Another approach, to address the lack of training data is represented by {\em multi-task learning}. This approach has been demonstrated effective for multi-lingual speech recognition, especially if the size of training data for each language is small \cite{miao2013,povey2014,huang2013,ghoshal2013,tang2016}. The reason of this is due to the fact that the shared hidden layers of the DNN used to estimate the emission probabilities of HMM states in a hybrid Automatic Speech Recognition (ASR) system \cite{hinton2012},  are language independent  if the DNN itself is trained on multi-lingual data. This DNN can be used to initialize a new one which can be  trained only with data of the target language to recognize. When the size of training data is small only a subset of the connection weights, usually those of the output layer,  are re-estimated.  This training procedure is often called {\em transfer learning}, to indicate the fact that an initial  set of learned parameters is transferred to the final acoustic model used by the ASR system.

In this work we address the problem of automatic speech recognition of children speaking a non-native language, specifically: ({\em a}) Italian students, speaking both English and German, and ({\em b}) German students speaking English.

It is known that non-native speakers articulate sounds very differently from native ones, because they try to use the phonology of their mother language, giving rise to two types of errors \cite{russell2007}: {\em mispronunciation}, when they aim to pronounce a wrong target, and {\em phonological interference} since they use their original set of phones. In the past several approaches have been proposed to take into account  the pronunciation errors of non-native speakers \cite{bouselmi2005,bouselmi2006}, spanning from the usage of non-native pronunciation lexicon \cite{Wang2003a,Wang2003b,Oh2006,strik2009,Steidl2004} to acoustic model adaptation using either  native data and non native data \cite{duan2017,li2016,lee2015,das2015}.

%\textcolor{red}{DANIELE: spiega DNN adaptation}
As previously mentioned, we use transfer learning to adapt the multi-lingual DNN trained on native data from Italian, German and English children. Basically, only the weights of the output layer of the network are updated, through back propagation, using data from non-native speakers of a given language while the weights of the lower layers are frozen and remain unchanged during adaptation.
In addition, we propose to use multi-lingual data even in the adaptation phase, that is to update the weights of the output layer of the original DNN with all available  non-native data.

The novelties of this work are: 
\begin{itemize}
\item
the application of both multi-task learning and transfer learning to the recognition of voices of children speaking in a foreign language; 
\item
the usage of (non-native) multi-lingual data for updating the weights of the original transferred DNN.
\end{itemize}

Experimental results reported in the paper show that: {\em (a)} the usage of the multi-lingual DNN gives performance, on native data, which are similar to those achieved with mono-lingual networks, i.e.\ trained only with data from a single language; {\em (b)} the usage of the multi-lingual DNN provides performance on non-native data significantly better than those obtained with mono-lingual DNNs; {\em (c)} the usage of multi-lingual data in the adaptation process further increases the performance on non-native data.

This paper is organized as follows: Section~\ref{sec:speechdatabases} describes the experimental data used for testing the approaches proposed in this work, Section~\ref{sec:ASR} gives details of acoustic models, language models and IPA based lexicon used in the multi-lingual ASR system employed  in this work;  section~\ref{sec:expe} reports experiments and related results. Finally, Section~\ref{sec:concl} concludes the paper, presenting directions for future work.

%
% Diego
%It  is  well  known  that  spectral and  temporal  characteristics  ofchildren's   speech   are  highly   influenced   by  the   anatomical,physiological and  developmental changes that occur  during the growth and     are     hence     different     from    those     of     adult speakers~\cite{KeFo80,LeePotNar99,HubStaCurAshJoh99}.             These differences  are  attributed mainly  to  anatomical and  morphological differences  in  the vocal-tract  geometry~\cite{FitGie99,MugHir2012}, less precise control of the articulators and a less refined ability to control         suprasegmental         aspects         such         as prosody~\cite{MclBle2003,Rus2007,Ham2014}. When an ASR system trained  on adults' speech is employed to recognize children's speech,  performance decreases drastically,  especially for younger children \cite{WilJac96,DasNixPic98,LiRus01,GiuGer03,PotNar03,GerGiuBru07,GerGiuBru09,liao2015,serizel2016}.
%%%%%%%%%%%

\section{SPEECH DATA}
\label{sec:speechdatabases}

In this work we exploited speech data collected within the European funded project PF-Star (2002-2004). During the PF-Star project, noticeable amount of speech data were collected from
English, German, Italian and Swedish children~\cite{batliner2005}; for one of the Italian corpora see also ~\cite{gerosa2007}. For the purposes of this work, children's speech pronounced by English, German and Italian students in the three languages were considered, as shown in Table~\ref{tab:corpora1}. 
As already mentioned, data from native speakers were used to train Acoustic Models (AMs), while test was carried out on both native and non-native data sets; all the children are in the age range 9-10. In addition, a non-native adaptation set was used in transfer learning and, finally, performance, measured in terms of Word Error Rate (WER), was computed on both native and non-native evaluation data sets.

% three databases of native English, German and Italian children,
% as well as three databases of non-native English and German speech, 
% collected from German and Italian children, were used.

\begin{table}[bht]
\begin{center}
\begin{tabular}{lccc}  \hline
speakers $\backslash$ language & Italian      & German       & English      \\ \hline
 Italian           & train + eval & ada + eval   & ada + eval   \\
 German            & --           & train + eval & ada + eval   \\ 
 English           & --           & --           & train + eval \\ \hline
\end{tabular}
\caption{Native and non-native children's speech corpora used in the paper.}
\label{tab:corpora1}
\end{center}
\end{table}

\begin{table}[bht]
\begin{center}
\begin{tabular}{rlcccc}      \hline
\multicolumn{2}{c}{number of}  & language & total     &  running & lexicon \\ 
\multicolumn{2}{c}{speakers}   & spoken   & duration  &  words   & size    \\ \hline
\multicolumn{6}{c}{mono-lingual native training corpora} \\
115&Italian & Italian  & 07:15:53  &  49233   & 9519   \\
168&German  & German   & 07:45:31  &  49326   & 7451   \\
 70&English & English  & 06:04:49  &  26873   & 1267   \\ \hline
 \multicolumn{6}{c}{mono-lingual native eval corpora} \\
 42&Italian & Italian  & 02:37:07 & 17936    & 5042    \\ 
 11&German  & German   & 01:21:18 &  7859    & 1948    \\
 30&English & English  & 01:40:38 &  9224    & 1036    \\ \hline
\end{tabular}
\caption{Details for mono-lingual, native, training and eval corpora.}
\label{tab:corporatraineval}
\end{center}
\end{table}

Table~\ref{tab:corporatraineval} reports details about the mono-lingual training and eval data - in all cases, native children's speech - in terms of number of speakers, duration, number of running words and lexicon size.
Table~\ref{tab:corporadeveval} presents some statistics about the non-native speech data. In particular, Italian children produced both English and German speech, while German children produced English speech. In all cases, the speech data was split into {\em ada} (used to perform transfer learning) and {\em eval} data (for evaluation purposes only). Overlapping among training, ada and eval speakers never occurs.
%A detailed description of the data collected during the PF-Star project is given in the papers~\cite{gerosa2004,russell2007}.

\begin{table}[bht]
\begin{center}
\begin{tabular}{rlcccc}      \hline
\multicolumn{2}{c}{number of}  & language & total     &  running & lexicon \\
\multicolumn{2}{c}{speakers}   & spoken   & duration  &  words   & size    \\ \hline
\multicolumn{5}{c}{non-native ada corpora}					\\ 
  9 &Italian & German  & 00:29:54  &  2575    &  438    \\ 
 21 &Italian & English & 00:58:24  &  2753    &  390    \\ 
 42 &German  & English & 00:30:19  &  3081    &  597    \\ \hline
\multicolumn{5}{c}{non-native eval corpora}					\\ 
 13 &Italian & German  & 00:46:14  &  3769    &  474    \\ 
 27 &Italian & English & 01:16:29  &  3632    &  444    \\ 
 52 &German  & English & 00:43:14  &  4440    &  630    \\ \hline
\end{tabular}
\caption{Details for non-native ada and eval corpora.}
\label{tab:corporadeveval}
\end{center}
\end{table}

\COMMENT{
\begin{table}[bht]
\begin{center}
\begin{tabular}{cccc} \hline
source/target language	&  Italian & German & English   \\ \hline
\multicolumn{4}{c}{test}					\\ \hline
Italian 				&  157  	& 46 	& 76 \\
German  				&  -   		& 81	& 43  \\
English 				&  -  		& -	    & 100 \\ \hline
\multicolumn{4}{c}{ada}	 \\ \hline
Italian 				&  -  		& 29 	& 57 \\
German  				&  -   		& -		&  30 \\ \hline
\end{tabular}
\label{tab:setssize}
\caption{Size (in minutes) of datasets considered in the experiments.}
\end{center}
\end{table}
}

\section{ASR system}
\label{sec:ASR}

As mentioned in the Introduction a multi-lingual DNN was first trained on data of native speakers. 

\subsection{Multi-lingual DNN}
%\textcolor{red}{MARCO: Short description of the multi-lingual DNN. Feed forward, usage of kaldi, do not give too much details on acoustic features, HMMs, ... . Give some details on alignments of non-native data}

The ASR system is based on the KALDI open source software toolkit \cite{Povey_ASRU2011}. 
The baseline acoustic model is build following the Karel's DNN recipe~\cite{karel2011}: the preliminary HMM is trained on the usual 13 mel-frequency cepstral coefficients  (MFCCs), which are then mean/variance normalized; fMLLR-transformed coefficients are then estimated and used as input features for the DNN. The learning procedure features layer-wise pre-training based on Restricted Boltzmann Machines, per-frame cross-entropy training and sequence-discriminative training (lattice framework and State Minimum Bayes Risk criterion).
Besides the mono-lingual DNNs trained on native German, Italian, English speech, a multi-lingual model is derived from a shared lexicon (see Section \ref{subsec:lexicon} for details): Table~\ref{tab:multimodels} Table 4 reports the number of phonetic units used by the mono-lingual lexica as well as by the multi-lingual lexicon; it also reports the size of the output layer of the DNNs trained.for mono-lingual and multi-lingual speech recognition.

\begin{table}[bht]
\begin{center}
\begin{tabular}{llccc}
    \hline
  		&  units & dnn output   \\ \hline
Italian &   28   & 1679 \\
German  &   45   & 1592 \\
English &   43   & 1526 \\ \hline
multi 	& 	67	 & 1632 \\ \hline
\end{tabular}
\caption{Number of phonetic units and size of the output layer of the DNNs trained for mono- and multi-lingual DNNs.}
\label{tab:multimodels}
\end{center}
\end{table}

\subsection{Language Models}

Since the focus of this paper is on acoustic modeling, we did not cope with  Out-Of-Vocabulary (OOV) words issues, that would have
complicated the analysis of results.  Also, in order to avoid to use different LMs for native and non-native corpora, we decided to build only a single LM for each language.
For this reason, the lexicon of each language has to contain at least all of the words included in the native eval, non-native ada and eval sets.
Given that the lexicon size is quite high (1289 to 5042 words), we
decided to use a bigram LM, with Witten-Bell smoothing, that assures a
reasonable perplexity over the ada + eval data, as reported in
Table~\ref{tab:lmdata}.

\begin{table}[bht]
\begin{center}
\begin{tabular}{lccccc}      \hline
language   & lexicon & running &  2-grams  &  PP  & OOV   \\  
           & size    & words   &           &      &       \\ \hline 
Italian    & 5042    & 17936   &  13854   & 38.2 & 0.0\% \\
German     & 2194    & 14203   &   6959   & 20.6 & 0.0\% \\
English    & 1289    & 23130   &   4983   & 25.0 & 0.0\% \\ \hline
\end{tabular}
\caption{Text data used to build the three LMs.}
\label{tab:lmdata}
\end{center}
\end{table}

\subsection{Lexicon}
\label{subsec:lexicon}

Concerning the lexicon, we have at our disposal grapheme to phoneme
converter for the three languages, that were used in the past to build
mono-lingual ASR systems.  For this work, we decided to convert all the mono-lingual phones in IPA format, shown here as ASCII sequences.
Of course, some of the choices we did (for instance, we replaced geminate consonants with simple ones for the Italian lexicon) are questionable and some of them could be revised in the future.
Table~\ref{tab:phones} contains the list of all the 67 phones resulting from the merging of the three lexica.  Of these, 18 phones are common for all three languages, 13 are common to only 2 languages (9 de+en, 2 it+de, 2 it+en), and 36 are present in one language only (16 de, 14 en, 6 it).

\begin{table}[bht]
\begin{center}
\begin{footnotesize}
\begin{tabular}{|lccc|lccc|lccc|}      \hline
A"   &    & de &    &  OW   &    &    & en &  j    & it & de & en \\ \hline
AA   &    &    & en &  OY   &    & de & en &  k    & it & de & en \\ \hline
AE   &    &    & en &  O    &    & de & en &  l    & it & de & en \\ \hline
AH   &    &    & en &  R    &    & de &    &  m    & it & de & en \\ \hline
AI   &    & de & en &  S    & it & de & en &  n    & it & de & en \\ \hline
AU   &    & de & en &  TH   &    &    & en &  o:   &    & de &    \\ \hline
AX   &    &    & en &  U"   &    & de &    &  o    & it &    &    \\ \hline
C    &    & de &    &  UA   &    &    & en &  p    & it & de & en \\ \hline
DH   &    &    & en &  U    &    & de & en &  pf   &    & de &    \\ \hline
E@   &    & de &    &  Z    &    &    & en &  r    & it & de & en \\ \hline
EA   &    &    & en &  a:   &    & de &    &  s    & it & de & en \\ \hline
ER6  &    & de &    &  a    & it & de &    &  tS   & it & de & en \\ \hline
ER   &    &    & en &  b    & it & de & en &  t    & it & de & en \\ \hline
EY   &    &    & en &  dZ   & it & de & en &  ts   & it & de &    \\ \hline
E    &    & de & en &  d    & it & de & en &  u"   &    & de &    \\ \hline
IA   &    &    & en &  dz   & it &    &    &  u:   &    & de &    \\ \hline
I    &    & de & en &  e:   &    & de &    &  u    & it &    & en \\ \hline
J    & it &    &    &  e    & it &    &    &  v    & it & de & en \\ \hline
L    & it &    &    &  f    & it & de & en &  w    & it &    & en \\ \hline
NG   &    & de & en &  g    & it & de & en &  x    &    & de &    \\ \hline
O"2  &    & de &    &  h    &    & de & en &  z    & it & de & en \\ \hline
O"9  &    & de &    &  i:   &    & de &    &       &    &    &    \\ \hline
OH   &    &    & en &  i    & it &    &    &       &    &    &    \\ \hline
\end{tabular}
\end{footnotesize}
\caption{List of IPA-like (expressed in ASCII characters) phones used for the three languages.}
\label{tab:phones}
\end{center}
\end{table}

\section{Experiments and Results}
\label{sec:expe}
%\textcolor{red}{MARCO: describes experiments, include the tables}

Table~\ref{tab:baselinesystem} reports the WER results obtained using the four reference acoustic models: the three models, trained on clean data from native children speaking Italian, German and English, perform slightly better than the multi-lingual models trained on the three corpora using the multi-lingual lexicon.
Nevertheless, the multi-lingual system shows more robustness against non-native speech: the off-diagonals WERs, that represent the performance for young students speaking a foreign language, are significantly lower when the target language is English (39.8\% against 53.2 and 31.7 against 40.4) whilst it similar in case of German (i.e.\ 18.5\% against 17.4\%).  

\begin{table}[bht]
\begin{center}
\begin{tabular}{llccc}
    \hline
%eval  					&  Italian & German & English   \\ \hline
speakers $\backslash$ language   	&  Italian & German & English   \\ \hline
%models 										\\ \hline
\multicolumn{4}{c}{mono-lingual AMs} 	\\
Italian 				&  2.1     	& 17.4 	& 53.2 \\
German  				&  -   		& 7.3	& 40.4  \\
English 				&  -  		& -	    & 8.0 \\ \hline 
%\multirow{3}{*}{multi} 	& 2.2 	    & 18.5  & 39.8 \\ 
\multicolumn{4}{c}{multi-lingual AM} \\
Italian						& 2.2 	    & 18.5  & 39.8 \\ 
German						& -		 	& 7.9	& 31.7 \\ 
English            			& -		 	& - 	& 10.4	\\ \hline
\end{tabular}
\caption{Results (WERs) using mono and multi-lingual acoustic models; the off-diagonals numbers represent WERs obtained on non-native speech.}
\label{tab:baselinesystem}
\end{center}
\end{table}

The forthcoming experiment shows the effectiveness of transfer learning in this applicative context: the baseline DNNs are adapted to non-native speech using limited data from the target domain. The adaptation sets comprise data of Italian students reading German and English sentences and German students reading English text.
The adaptation is implemented keeping fixed all the layers of the DNNs except the last one; few additional learning iterations are then performed using the adaptation material.
We evaluated three modalities related to transfer learning: 
{\em m1)} the mono-lingual model is adapted to non-native speakers using adaptation data of the single target language;
{\em m2)} similarly, the multi-lingual model is adapted to a single target language;
{\em m3)} the multi-lingual model is adapted to multi-lingual non-native speech using at the same time the three adaptation sets (i.e. English and German data coming from Italian speakers and English data coming from German speakers). 

The results presented in Table~\ref{tab:adapted_systems} demonstrate that the multi-lingual system is capable to better cope with the acoustic mismatch introduced by non-native speakers. Indeed, also in this case, the multi-lingual system exhibits a better behavior, allowing to produce more effective models for non-native speech: the WERs related to Italian students speaking in both German and English decrease from 11.1\% to 9.6\% and from 16.0\% to  15.4\%, respectively; larger gain is observed in case of Germans speaking in English (WER from 18.3\% to 15.2\%).

\begin{table}[bht]
\begin{center}
\begin{tabular}{llccc}
    \hline
speakers $\backslash$ language			& German & English   \\ \hline
\multicolumn{3}{c}{adapted mono-lingual AMs ({\em m1})} 	\\
Italian 		&  11.1 	& 16.0 \\
German  		&  -		& 18.3  \\  \hline
\multicolumn{3}{c}{adapted multi-lingual AM ({\em m2})} 	\\
Italian		 	& {\bf 9.6} & 15.4 \\ 
German			& -	& 15.2 \\   \hline
\multicolumn{3}{c}{adapted multi-lingual AM ({\em m3})} 	\\
Italian	& 10.4 & 15.0 \\ 
German			&	- &	15.1 \\ \hline
\end{tabular}
\caption{WERs obtained with mono- and multi-lingual acoustic models adapted to non-native speech using the three modalities {\em m1,m2,m3}.}
\label{tab:adapted_systems}
\end{center}
\end{table}

%\textcolor{red}{metto anche questi risultati, vediamo poi se vale la pena o sono poco significativi}

The next experiment (see Table~\ref{tab:adapted_systems2}) investigates the case where, starting from the multi-lingual model, we adapt from Italian: adaptation sets from 
Italian students speaking German and English are merged. Similarly, the English set merges data of German and Italian students speaking English. In the first case, we use the adaptation sentences coming from speakers with the same mother tongue (i.e., Italian). Vice versa, in the second case, the adaptation set is defined in terms of the language spoken by the students, regardless of their original mother language. 
Also in this case, the gain with respect to the mono-lingual case is noticeable.  
%For example, the result related to Italian students speaking English moves from 15.4\% (i.e., adapting using only the corresponding dev set) 
Moreover, this combination can lead to the best results (highlighted in bold) for two cases; of course, as expected, the pairs with no adaptation data (German-to-English and Italian-to-German, respectively) produce lower results.

\begin{table}[bht]
\begin{center}
\begin{tabular}{llccc}
    \hline
speakers $\backslash$ language 			& German & English   \\ \hline
\multicolumn{3}{c}{adapted multi-lingual AM with Italian speakers} 	\\
Italian 	& 10.3 & {\bf 14.2} \\ 
German		&	- &	19.8 \\ \hline 
\multicolumn{3}{c}{adapted multi-lingual AM with English utterances } \\
Italian 	& 16.7 & 14.8 \\ 
German		&	- &	{\bf 15.0}\\ \hline
            
\end{tabular}
\caption{WERs related to experiments exploring modalities in which speakers are merged according to source or target language (i.e. Italian native speakers and students speaking in English, respectively).}
\label{tab:adapted_systems2}
\end{center}
\end{table}
As a consequence, we conclude that transfer-learning in the context of children non-native speech, where usually a limited amount of data is available for training purposes, can successfully be applied and mitigate the acoustic mismatch.
Moreover, it seems evident that the hidden layers of the multi-lingual DNN are able to build a more general representation of the phonetic space and this turns out to be suitable for the adaptation to non-native speakers.

\section{Conclusions}
\label{sec:concl}

In this work we have investigated the application of transfer learning for adapting a multi-lingual DNN, trained on native speech from three languages (Italian, German, English), to non-native data. The approach is implemented updating the output layer of the DNN with small adaptation sets; the experimental results confirm the validity of this technique and show the positive effect of a multi-language model to compensate the pronunciation differences of a non-native speaker.   
%Further performance improvements were achieved using multi-lingual data even in the adaptation step.

As future work we plan to further address non-native acoustic modeling, experimenting other types of acoustic features and exploiting some a-priori knowledge about phonetics of the first and foreign languages. In particular, it seems promising the investigation of alternative lexicons that take into account the possible pronunciation variations introduced by non-native students.

%\vfill
%\pagebreak

%\vfill\pagebreak

%\section{REFERENCES}
%\label{sec:refs}

\vfill \pagebreak
\ninept
\bibliographystyle{IEEEbib}

\bibliography{strings,refs}

\end{document}